\documentclass[journal]{IEEEtran}
\usepackage{amssymb,amsmath,algorithmic,algorithm}
\usepackage{graphicx,url}

\ifCLASSINFOpdf
\else
\fi
\ifCLASSOPTIONcompsoc
  \usepackage[caption=false,font=normalsize,labelfont=sf,textfont=sf]{subfig}
\else
  \usepackage[caption=false,font=footnotesize]{subfig}
\begin{document}
%
\title{Doubly Contrastive Deep Clustering}
%
%
%

\author{Zhiyuan Dang, Cheng Deng, Xu Yang, Heng Huang 
	\IEEEcompsocitemizethanks{\IEEEcompsocthanksitem Z.Y. Dang, C. Deng and X. Yang are with School of Electronic Engineering, Xidian University, Xi'an, China (email: zydang@stu.xidian.edu.cn, chdeng@mail.xidian.edu.cn and xyang\_01@stu.xidian.edu.cn).
	\IEEEcompsocthanksitem H. Huang is with  Department of Electrical \& Computer Engineering, University of Pittsburgh, USA and with JD Finance America Corporation (e-mail:heng.huang@pitt.edu).}
	\thanks{H. Huang and C. Deng are the corresponding authors.}}

%
%

\markboth{Journal of \LaTeX\ Class Files,~Vol.~14, No.~8, August~2015}%
{Shell \MakeLowercase{\textit{et al.}}: Bare Demo of IEEEtran.cls for IEEE Journals}
%



\maketitle


\begin{abstract}
	Deep clustering successfully provides more effective features than conventional ones and thus becomes an important technique in current unsupervised learning. However, most deep clustering methods ignore the vital positive and negative pairs introduced by data augmentation and further the significance of contrastive learning, which leads to suboptimal performance. In this paper, we present a novel Doubly Contrastive Deep Clustering (DCDC) framework, which constructs contrastive loss over both sample and class views to obtain more discriminative features and competitive results. Specifically, for the sample view, we set the class distribution of the original sample and its augmented version as positive sample pairs and set one of the other augmented samples as negative sample pairs. After that, we can adopt the sample-wise contrastive loss to pull positive sample pairs together and push negative sample pairs apart. Similarly, for the class view, we build the positive and negative pairs from the sample distribution of the class. In this way, two contrastive losses successfully constrain the clustering results of mini-batch samples in both sample and class level. Extensive experimental results on six benchmark datasets demonstrate the superiority of our proposed model against state-of-the-art methods. Particularly in the challenging dataset Tiny-ImageNet, our method leads 5.6\% against the latest comparison method. Our code will be available at \url{https://github.com/ZhiyuanDang/DCDC}.
\end{abstract}


\section{Introduction}
\IEEEPARstart{S}{ince} label obtainment becomes more expensive, many researchers gradually turn to work on unsupervised learning. As an important branch of unsupervised learning, clustering methods have attracted more attention, which aim to find a good partition that divide similar samples into the same cluster while dissimilar ones into different clusters. 

Recent clustering methods (such as K-Means \cite{macqueen1967some}, Spectral Clustering \cite{zelnik2005self}, Nonnegative Matrix Factorization Clustering \cite{cai2009locality}) have been widely used in various tasks. However, these methods only focus on local pixel-level information, ignoring more higher-level and semantic one which limits their performance a lot. By virtue of the powerful deep learning \cite{lecun2015deep,ijcai2020-77,wei2019adversarial}, deep clustering methods are proposed to non-linearly transform the original data into latent space for successfully provide more effective features and promising performance. Even so, deep clustering methods still perform bad in more challenging datasets such as ImageNet \cite{russakovsky2015imagenet} and some subsets of ImageNet (such as STL-10 \cite{coates2011analysis}, Tiny-ImageNet \cite{le2015tiny} and so on). Therefore, how to further improve the performance in such these datasets is still an open question.

\begin{figure}[!t]
	\centering
	\subfloat[Sample View]{
		\centering
		\includegraphics[width=0.4\textwidth]{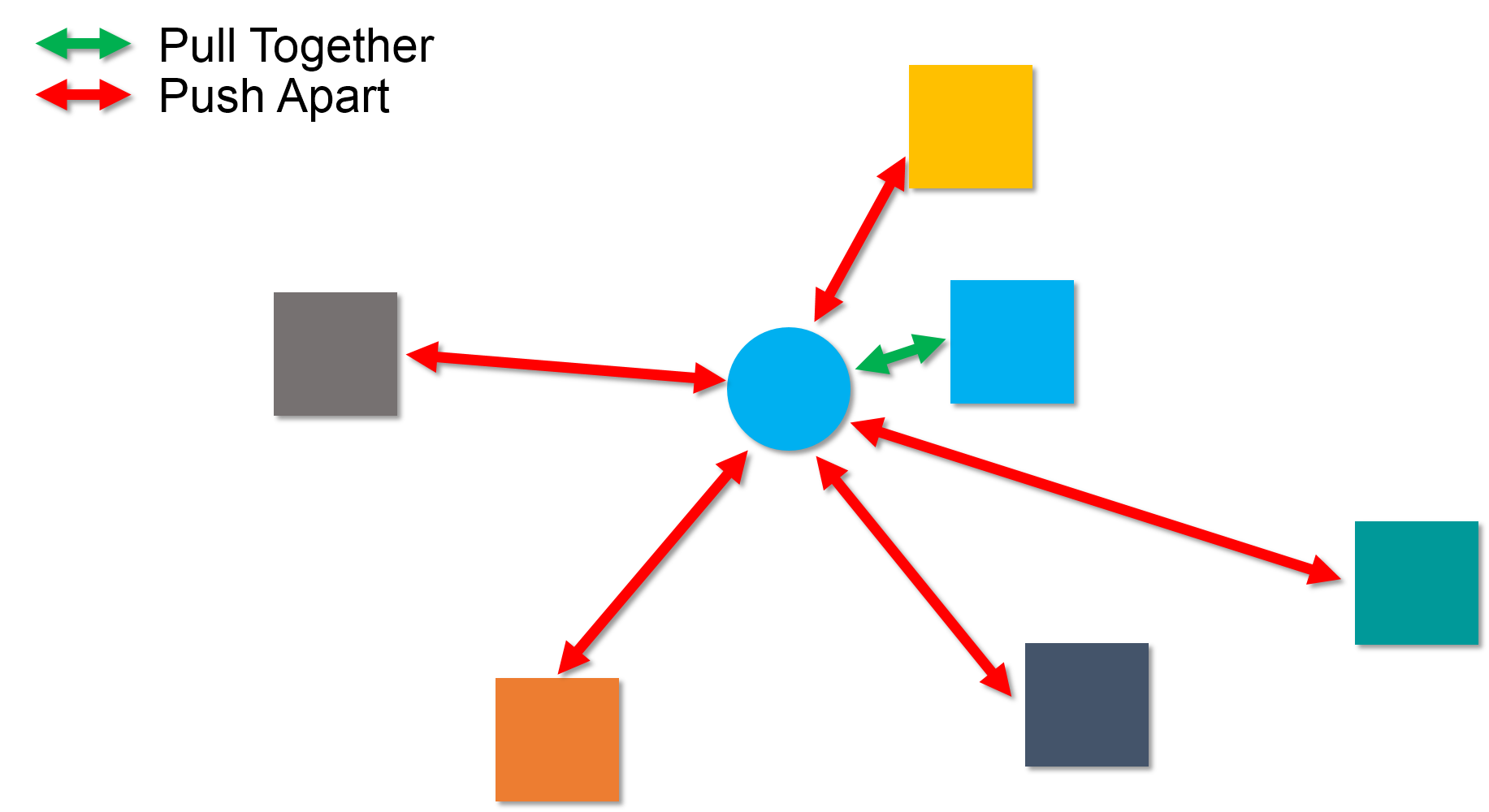}
		\label{illustration_sample}
	}\\
	\subfloat[Class View]{
		\centering
		\includegraphics[width=0.4\textwidth]{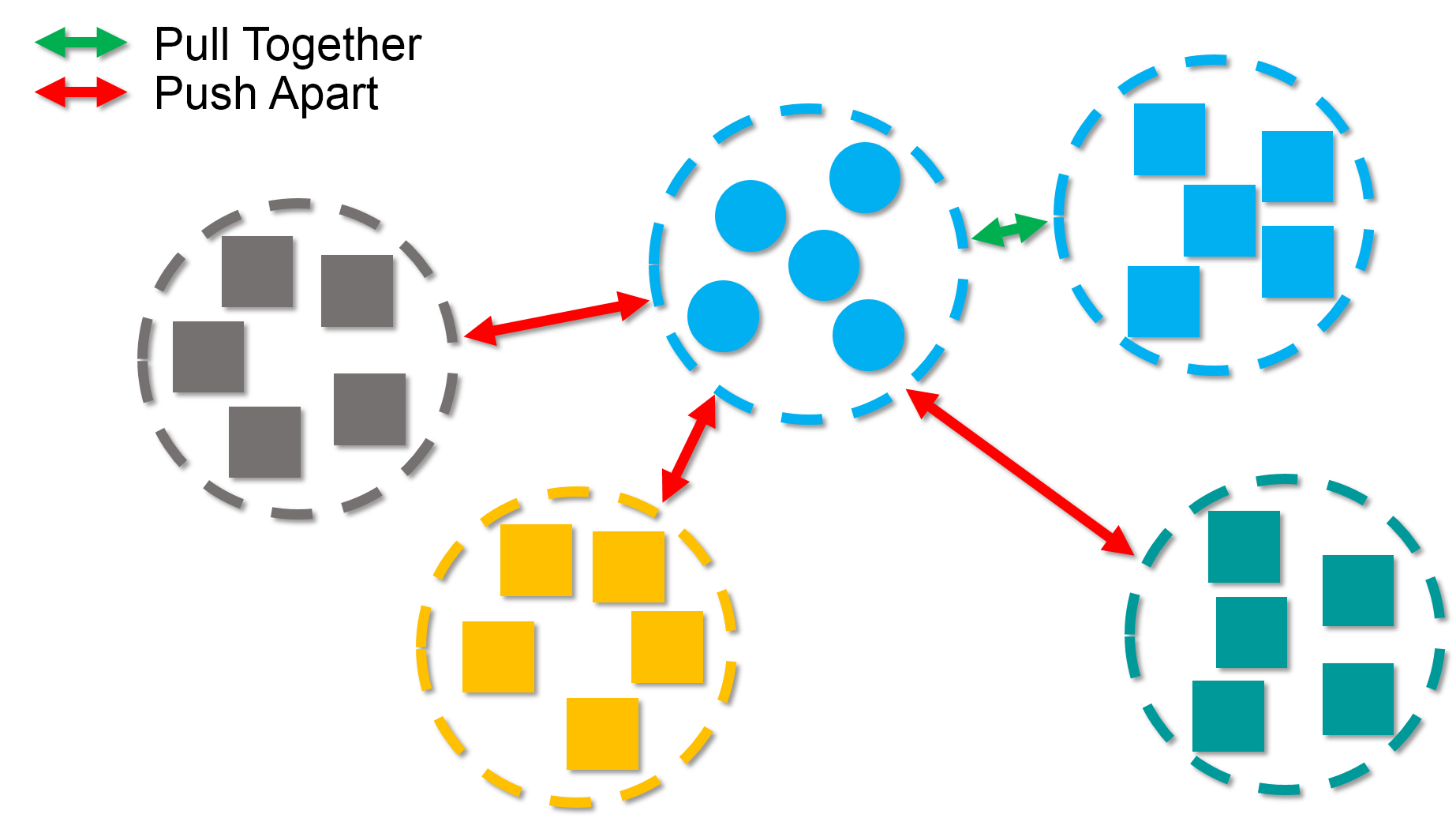}
		\label{illustration_class}
	}
	\caption{Illustration of our idea. Do contrastive learning from two views: sample and class view. The goal of sample view is to pull positive sample pair together and push negative sample pairs apart. Class view intends to pull positive class pair together and push negative class pairs apart.}
	\label{illustration}
\end{figure}

Current researchers address this problem from two aspects. On the one hand, some methods \cite{chang2017deep,chang2019deep,wu2019deep} have been proposed to exploit the inter-samples relationship from the pairwise distance between the latest sample features and thus update the model. However, since the inconsistent estimations in the neighborhoods during training, these approaches may suffer from more severe error-propagation issues which lead to suboptimal solution. 

On the other hand, since the useful data augmentation technique to enhance the robustness of the model, some methods \cite{haeusser2018associative,ji2019invariant} were proposed to maximize the mutual information (MI) between the class distribution of original samples and their augmented ones. Although coursing the clustering problem from sample view (similar to Figure~\ref{illustration_sample}), they ignore the vital positive and negative class pairs introduced by data augmentation, i.e., think from class view (as shown in Figure~\ref{illustration_class}). With this insight, the latest method \cite{huang2020deep} maximizes the MI between the sample distribution of original classes and their augmented ones. However, among from above methods, they only pursue improving clustering performance from one view, usually miss another one, which result in inferior performance.

Considering deal with the problem from both sample and class views, therefore, in this paper, we propose a novel deep clustering framework based on two contrastive losses, named Doubly Contrastive Deep Clustering (DCDC). Specifically, according to data augmentation technique, since an ideal image representation should be invariant to the geometry transformation, we can view the class distribution of the original sample and its augmented version as positive sample pair and set the one of other augmented samples as negative pairs. For positive pair, we use sample-wise contrastive loss to pull them together; as for negative pairs, we push them apart. That also holds for class-wise contrastive loss: we can view the sample distribution of original class and its augmented one as positive class pair and set the one of other augmented classes as negative pairs. As shown in Figure~\ref{illustration}, our DCDC considers the clustering results of the classification head of the backbone network from both sample and class views. Based this novel doubly contrastive loss, our method can obtain more discriminative features and then achieve better clustering results.

Our major contributions can be summarized as follows:
\begin{itemize}
	\item We propose a novel deep clustering framework based two contrastive losses, called Doubly Contrastive Deep Clustering (DCDC). Different from previous methods, DCDC considers improving clustering performance from both sample and class views. To best of our knowledge, it is the first contrastive learning based deep clustering method.
	\item We provide the desired affinity matrix of both sample and class views, after optimizing doubly contrastive loss. As shown in Figures~\ref{framework}, \ref{sample_affinity_tiny} and \ref{class_affinity_tiny}, desired sample-wise affinity matrix $M$ is block-diagonal structure (if data are sort by class) and desired class-wise affinity $N$ should be diagonal. Besides that, we also conduct more ablation studies about over-clustering, sample repeat and batch size to figure out the useful techniques or tricks for contrastive based deep clustering. 
	\item Extensive experimental results on six benchmark datasets demonstrate the superiority of our DCDC against other state-of-the-art methods. Particularly in the challenging dataset Tiny-ImageNet, our DCDC leads 5.6\% against the latest comparison method.
\end{itemize}

\section{Related Work}
\subsection{Deep Clustering}
According to the data training strategy, current deep clustering approaches \cite{yang2019deep,yang2020adversarial,dang2020multi} could be roughly divided into two categories: The first one (such as DEC \cite{xie2016unsupervised}, JULE \cite{yang2016joint}, DAC \cite{chang2017deep},  DeepCluster \cite{caron2018deep}, DDC \cite{chang2019deep}, DCCM \cite{wu2019deep}) usually iteratively evaluate the clustering assignment from the up-to-date model and supervise the network training processes by the estimated information; The second one (such as ADC \cite{haeusser2018associative}, IIC \cite{ji2019invariant}, PICA \cite{huang2020deep}) simultaneously learn both the feature representation and clustering assignment at the same time without explicit phases of clustering.

For alternate training based works, DEC \cite{xie2016unsupervised} is a typical method that initializes clustering centroids by applying K-Means \cite{lloyd1982least} on pre-trained image features and then fine-tunes the model to learn from the confident clustering assignments to sharpen the resulted prediction distribution. JULE \cite{yang2016joint} jointly optimizes the CNN in a recurrent manner, where merging operations of agglomerative clustering are conducted in the forward pass and representation learning is performed in the backward pass. DAC \cite{chang2017deep}, DDC \cite{chang2019deep} and DCCM \cite{wu2019deep} alternately update the clustering assignment and inter-sample similarities during training. However, they are susceptible to the inconsistent estimations in the neighborhoods and thus suffer from severe error-propagation problem at training.

For current simultaneous training works, they more likes combing deep representation learning \cite{bengio2007greedy,hjelm2018learning} with conventional cluster analysis \cite{gowda1978agglomerative,lloyd1982least} or other pretext objectives. ADC \cite{haeusser2018associative} constructs an optimization objective to encourage consistent association cycles among clustering centroids and sample embeddings. For the latest works, they are mostly based on mutual information maximization of the clustering assignment. For example, IIC \cite{ji2019invariant} and IMSAT \cite{hu2017learning} are proposed to learn a clustering assignment by maximizing the mutual information between an image and its augmentations. Although the methods from this category alleviate the negative influence of inaccurate supervision from estimated information which brings from alternate training works, their objectives are usually more ambiguous than those of the alternate approaches as they can be met by multiple different separations. Since the vague connection between the training supervision and clustering objective, PICA \cite{huang2020deep} deals with this limitation by introducing a partition uncertainty index to quantify the global confidence of the clustering assignment so as to select the most semantically plausible separation. 
SCAN \cite{van2020learning} is firstly proposed to further improve clustering semantics by a two-step procedure that firstly learns semantic features and then adopts the obtained features as a prior in a deep clustering network.

\subsection{Contrastive Learning}
Contrastive losses \cite{hadsell2006dimensionality} measure the similarities of sample pairs in a latent space which transformed usually by DNN. Instead of matching an input data to a fixed target, in contrastive loss formulations the target can be various on-the-fly during training and can be defined in terms of the data representation computed by a network. Based on this novel loss, contrastive learning becomes a hot topic on recent unsupervised learning works \cite{wu2018unsupervised,hjelm2018learning,oord2018representation,henaff2019data,he2020momentum,chen2020simple}. 

According to different numbers of sample pairs, current contrastive learning methods could be roughly divided into three categories. The first one is end-to-end mechanism (such as \cite{hadsell2006dimensionality,oord2018representation,hjelm2018learning,henaff2019data,chen2020simple}) which is common and back-propagation based. It uses samples of the current mini-batch to construct sample pairs. But the sample pair number is coupled with the mini-batch size, limited by the GPU memory size which also is challenged by large mini-batch optimization \cite{goyal2017accurate}. The second one is memory-bank based mechanism \cite{wu2018unsupervised}. A memory bank consists of all the samples features in the total dataset. Since the samples in mini-batch are randomly sampled from the memory bank without back-propagation, therefore, this method supports large batch size. However, the sample features in the memory bank are updated asynchronously with current sample features of the DNN, and thus are sometime less consistent. The third one is latest momentum encoder based mechanism \cite{he2020momentum}. \cite{he2020momentum} proposed to encode samples on-the-fly by a momentum-updated encoder, and maintain a queue of sample features. Since the simple implementation and less class numbers, we adopt end-to-end mechanism to do our contrastive learning based deep clustering method.

\begin{figure*}[!h]
	\centering
	\includegraphics[width=\linewidth]{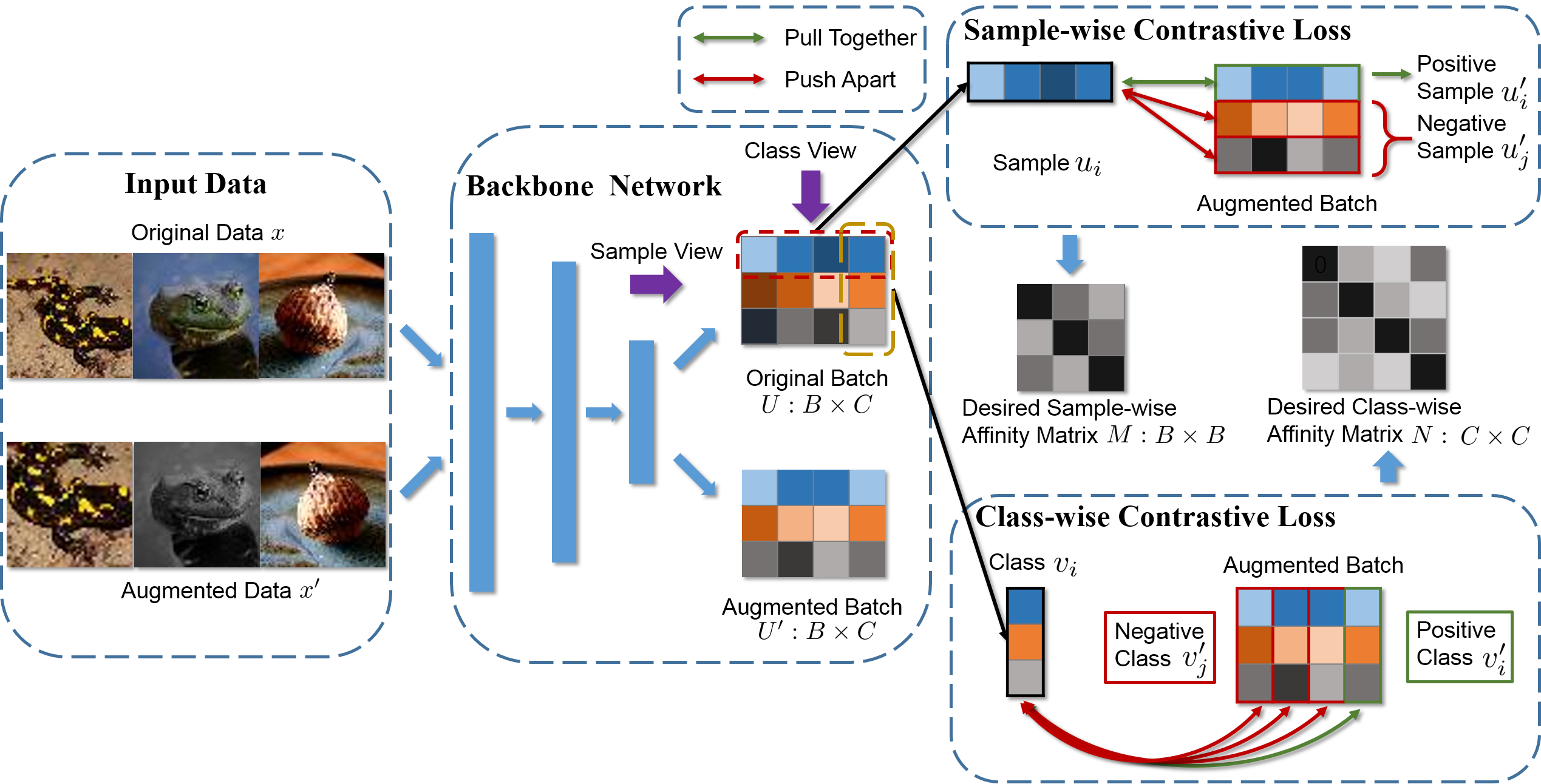}
	\caption{The framework of the proposed network: Doubly Contrastive Deep Clustering (DCDC). Firstly, we generate augmented data based on different transformations over original data and then input both of them to a backbone network. The network give the classification results of input data, i.e., original batch $U:B\times C$ and augmented batch $U':B\times C$, where $B$ is the batch size and $C$ is the class number. Considering original and augmented data is the same object in different transformation, we can view the class distribution of original sample and its corresponding augmented version as positive sample pair, and the one of other augmented samples as negative sample pairs, which also be applicable for class view. With these two insights, we naturally propose two contrastive losses from both sample and class views to obtain more discriminative results. After optimization, two affinity matrices based on the existing positive and negative pairs are expected to be shown as $M$ and $N$. Note that in practice, there are many samples belong to one class, therefore, $M$ will have block-diagonal structure (if data are sort by class) and $N$ still is diagonal structure (see ``Qualitative Study'' subsection of Experiment section for more details).}
	\label{framework}
\end{figure*}

\subsection{Mutual Information Estimation}
Aims to different tasks, we have various goal to optimize MI, i.e., MI is typically utilized as a criterion or a regularizer in loss functions, to enhance or constrain the dependence between variables. For example, contrastive representation learning (most methods in above subsection), generative models \cite{chen2016infogan} and information distillation \cite{ahn2019variational} are MI maximization problem. And disentangled representation learning \cite{chen2018isolating}, style transfer \cite{kazemi2018unsupervised}, domain adaptation \cite{gholami2020unsupervised} are MI minimization problem. 

However, MI is notoriously hard to estimate, since the computation requires closed forms of density functions and a tractable log-density ratio between the joint and marginal distributions \cite{cheng2020club}. In most machine learning tasks, only samples from the joint distribution are accessible. Therefore, most methods are sample-based MI estimations. For the MI maximization problem like deep clustering, researchers always prefer to maximize the lower bound of MI. There are many methods give the lower bound estimation of MI, such as  NWJ \cite{nguyen2010estimating}, MINE \cite{belghazi2018mine} and InfoNCE \cite{oord2018representation}. The results in \cite{hjelm2018learning} show that InfoNCE could obtain better performance and then widely used in latest contrastive representation learning methods (such as \cite{chen2020simple,he2020momentum}). Besides, \cite{tschannen2019mutual} declare that maximizing tighter bounds on MI may cause worse representations and provide the connection between deep metric learning and contrastive representation learning. Based on above analyses, we adopt InfoNCE loss as the based contrastive loss of our DCDC.

\section{Doubly Contrastive Deep Clustering}
In this section, we first review the InfoNCE loss \cite{oord2018representation} that is a form of contrastive loss function, and then provide the details of our DCDC loss. 

Let $x,x'\in\mathcal{X}$ be paired data samples from a joint probability distribution $p(x,x')$, i.e., $x$ and $x'$ could be different data include the same object. Since DNN always do data augmentation, we can denote original data sample as $x$, augmented data sample as $x'$ respectively. Our framework aims to learn a DNN mapping $\varPhi: \mathcal{X}\rightarrow\mathcal{Y}$ that fully discover the information between $x$ and $x'$, where $\mathcal{X}=\{(x_1,x'_1),(x_2,x'_2),\cdots,(x_N,x'_N)\}$ and $\mathcal{Y}=\{1,2,\cdots,C\}$, $N$ is the sample size, $C$ is the class number.

\subsection{InfoNCE Loss}
As mentioned above, since InfoNCE loss is the lower bound estimation of MI, we first present the definition of MI as follows:

\begin{equation}
	I(x,x')=\sum\limits_{x,x'}p(x,x')\log\frac{p(x'|x)}{p(x')}.\label{MI}
\end{equation}
Maximizing the MI between $x$ and $x'$, we can obtain more robust and discriminative deep features. Based on the analysis of \cite{oord2018representation}, directly modeling $p(x'|x)$ may not be optimal for learning more common information between $x$ and $x'$. Therefore, they preserve the density ratio between $x$ and $x'$ in Eq.~\ref{MI} as follow:

\begin{equation}
	f(x,x')\propto \frac{p(x'|x)}{p(x')},\label{proport}
\end{equation}where $\propto$ stands for `proportional to', the density ratio $f$ can be unnormalized and any positive real score. 

Since $p(x')$ and $p(x'|x)$ can not be estimated directly, we can adopt the samples drawn from these distributions. In this way, the Noise-Contrastive Estimation (NCE) \cite{gutmann2010noise} and Importance Sampling \cite{bengio2008adaptive} techniques that comparing target value with randomly sampled negative values can be considered. 

Based on NCE, given a dataset with $N$ samples containing one positive sample $x'_i$ of $x_i$ from conditional distribution $p(x'|x)$ and $N-1$ negative samples from the distribution $p(x')$, we give the InfoNCE loss as follow:

\begin{equation}
	\mathcal{L}_{InfoNCE}=-\mathbb{E}\left[ \log\frac{f(x_i,x'_i)}{\sum_{j=1}^Nf(x_i,x_j')}\right].\label{infoNCE} 
\end{equation}
The loss is actually a categorical cross-entropy loss of classifying the positive sample correctly with predicting $\frac{f}{\sum f}$. Therefore, the probability that sample $x_i'$ is drawn from conditional distribution $p(x'|x)$ rather than $p(x')$ can be written as follow:

\begin{align}
	p_{pos}&=\frac{p(x_i'|x_i)\prod_{m\neq i}p(x_m')}{\sum_{j=1}^Np(x_j'|x_i)\prod_{m\neq j}p(x_m')}\nonumber\\
	&=\frac{\frac{p(x_i'|x_i)}{p(x_i')}}{\sum_{j=1}^N\frac{p(x_j'|x_i)}{p(x_j')}}.\label{pos_prob}
\end{align}
The optimal value for $f(x,x')$ in Eq.~\ref{infoNCE} is proportional to $\frac{p(x'|x)}{p(x')}$. That is, optimizing the loss results in $f(x,x')$ estimate the density ratio in Eq.~\ref{proport}. Finally, \cite{oord2018representation} present InfoNCE is actually a lower bound of MI:

\begin{equation}
	I(x,x')\ge\log(N)-\mathcal{L}_{InfoNCE},\label{lower_bound}
\end{equation}which will become tighter with $N$ increasing. In addition, minimizing the InfoNCE loss will leads to the maximization of the lower bound of MI.

\subsection{DCDC Loss}
Following the InfoNCE loss, we proposed our Doubly Contrastive Deep Clustering (DCDC) loss. Figure~\ref{framework} shows the detailed framework of our DCDC network.  Firstly, for the same batch after classification head of backbone network, there are original clustering results $U\in\mathbb{R}^{B\times C}$ and augmented ones $U'\in\mathbb{R}^{B\times C}$. Our DCDC is constructed from both sample and class views of these two batches, which consists of sample-wise and class-wise contrastive losses. As follow, we provide the detailed descriptions about these two contrastive losses.
\subsubsection{Sample-wise Contrastive Loss:}
For the sample view, since samples $x$ and $x'$ are the same object after different transformation, these two samples should be classify into same class, i.e, their clustering results $u\in\mathbb{R}^{C\times 1}$ and $u'\in\mathbb{R}^{C\times 1}$ should have same class distribution. For one sample in each batch, it have one positive sample and $B-1$ negative samples. For easy implementation, we adopt the cosine similarity to measure the degree of the similarities between $u$ and $u'$, i.e.,

\begin{equation}\label{key}
	cos(u,u')=\frac{u^Tu'}{\|u\|_2\|u'\|_2},
\end{equation}where $\|\cdot\|_2$ is the $\ell_2$-normalization. Thus, two affinity matrices $M$ and $N$ in Figure~\ref{framework} can be obtained by $M=cos(UU'^T)$ and $N=cos(U^TU')$. As shown in \cite{wu2018unsupervised}, the dense ratio $f$ could be

\begin{equation}\label{f_sample}
	f(u,u')=\exp(cos(u,u')/\tau),
\end{equation} where $\tau>0$ is a temperature parameter. Since $u$ and $u'$ are normalized by softmax and $\ell_2$-norm, $cos(u,u')\in\left[0,1\right] $. We still view $u_i'$ as the positive sample of $u_i$, therefore, we obtain the sample-wise contrastive loss based Eq.~\ref{infoNCE} over batch:

\begin{equation}
	\mathcal{L}_{sample}=-\mathbb{E}\left[ \log\frac{\exp(cos(u_i,u'_i)/\tau)}{\sum_{j=1}^B\exp(cos(u_i,u'_j)/\tau)}\right].\label{sample_contra} 
\end{equation}
\begin{algorithm}[t]
	\renewcommand{\algorithmicrequire}{\textbf{Input:}}
	\renewcommand{\algorithmicensure}{\textbf{Output:}}
	\renewcommand{\algorithmicloop}{\textbf{where} $t<T:$ }
	\renewcommand{\algorithmicendloop}{\textbf{end while}}
	\caption{Training DCDC Network.}
	\begin{algorithmic}[1] 
		\REQUIRE Input data $\mathcal{X}$, training epochs $T$, iterations per epoch $I$, target class number $C$, temperature parameter $\tau$.
		\STATE Random initialization the backbone DNN weights $\varPhi$.
		\FOR{$t=1,\ldots,T$}
		\FOR{$i=1,\ldots,I$}
		\STATE Sampling a random mini-batch about images.
		\STATE Feeding the mini-batch images into the DNN model.
		\STATE Computing sample contrastive loss (Eq.\ref{sample_contra}) .
		\STATE Computing class contrastive loss (Eq.\ref{class_contra}).
		\STATE Minimizing the total loss (Eq.\ref{total_loss}) to update the DNN weights $\varPhi$.
		\ENDFOR
		\ENDFOR
		\ENSURE A DNN model with trained weights $\varPhi$.
	\end{algorithmic}
	\label{algorithm1}
\end{algorithm}
\subsubsection{Class-wise Contrastive Loss:}
For the class view, these two batches $U$ and $U'$ should have same sample distribution. Therefore, samples which be classified as the same class can be viewed as positive, i.e., $v\in\mathbb{R}^{B\times 1}$ and $v'\in\mathbb{R}^{B\times 1}$. Similar with Eq.\ref{f_sample}, the density ratio $f$ can be rewritten as:

\begin{equation}\label{f_class}
	f(v,v')=\exp(cos(v,v')/\tau),
\end{equation}the class-wise contrastive loss as follow:

\begin{equation}
	\mathcal{L}_{class}=-\mathbb{E}\left[ \log\frac{\exp(cos(v_i,v'_i)/\tau)}{\sum_{j=1}^C\exp(cos(v_i,v'_j)/\tau)}\right],\label{class_contra} 
\end{equation} which have one positive class and $C-1$ negative classes. Therefore, the total objective function of DCDC is formulated as: 

\begin{equation}
	\mathcal{L}=\mathcal{L}_{sample}+\mathcal{L}_{class}.\label{total_loss}
\end{equation}
\begin{table*}[ht]
	\centering
	\caption{The clustering performance (\%) on six challenging object image benchmarks. The best results are indicated in \textbf{Bold}.}
	\renewcommand\arraystretch{1.1}
	\setlength{\tabcolsep}{1mm}
	\begin{tabular}{c|ccc|ccc|ccc|ccc|ccc|ccc}
			\hline
			\multicolumn{1}{c|}{Datasets} & \multicolumn{3}{c|}{CIFAR-10} & \multicolumn{3}{c|}{CIFAR-100} & \multicolumn{3}{c|}{STL-10} & \multicolumn{3}{c|}{ImageNet-10} & \multicolumn{3}{c|}{ImageNet-Dogs} & \multicolumn{3}{c}{Tiny-ImageNet} \\
			\hline
			Metrics & NMI & ACC & ARI & NMI & ACC & ARI & NMI & ACC & ARI & NMI &  ACC& ARI & NMI & ACC & ARI & NMI & ACC & ARI\\ \hline                                               \hline
			K-Means & 0.087 & 0.229 & 0.049 & 0.084 & 0.130 & 0.028 & 0.125 & 0.192 & 0.061 & 0.119 & 0.241 & 0.057 & 0.055 & 0.105 & 0.020 & 0.065 & 0.025 & 0.005\\
			\hline
			SC & 0.103 & 0.247 & 0.085 & 0.090 & 0.136 & 0.022 & 0.098 & 0.159 & 0.048 & 0.151 & 0.274 & 0.076 & 0.038 & 0.111 & 0.013 & 0.063 & 0.022 & 0.004\\
			\hline
			AC & 0.105 & 0.228 & 0.065 & 0.098 & 0.138 & 0.034 & 0.239 & 0.332 & 0.140 & 0.138 & 0.242 & 0.067 & 0.037 & 0.139 & 0.021 & 0.069 & 0.027 & 0.005\\
			\hline
			NMF & 0.081 & 0.190 & 0.034 & 0.079 & 0.118 & 0.026 & 0.096 & 0.180 & 0.046 & 0.132 & 0.230 & 0.065 & 0.044 & 0.118 & 0.016 & 0.072 & 0.029 & 0.005\\
			\hline
			AE & 0.239 & 0.314 & 0.169 & 0.100 & 0.165 & 0.048 & 0.250 & 0.303 & 0.161 & 0.210 & 0.317 & 0.152 & 0.104 & 0.185 & 0.073 & 0.131 & 0.041 & 0.007\\  
			\hline                                        DAE & 0.251 & 0.297 & 0.163 & 0.111 & 0.151 & 0.046 & 0.224 & 0.302 & 0.152 & 0.206 & 0.304 & 0.138 & 0.104 & 0.190 & 0.078 & 0.127 & 0.039 & 0.007\\
			\hline                                        DCGAN & 0.265 & 0.315 & 0.176 & 0.120 & 0.151 & 0.045 & 0.210 & 0.298 & 0.139 & 0.225 & 0.346 & 0.157 & 0.121 & 0.174 & 0.078 & 0.135 & 0.041 & 0.007\\
			\hline                                        DeCNN & 0.240 & 0.282 & 0.174 & 0.092 & 0.133 & 0.038 & 0.227 & 0.299 & 0.162 & 0.186 & 0.313 & 0.142 & 0.098 & 0.175 & 0.073 & 0.111 & 0.035 & 0.006\\
			\hline                                        VAE & 0.245 & 0.291 & 0.167 & 0.108 & 0.152 & 0.040 & 0.200 & 0.282 & 0.146 & 0.193 & 0.334 & 0.168 & 0.107 & 0.179 & 0.079 & 0.113 & 0.036 & 0.006\\
			\hline                                        JULE & 0.192 & 0.272 & 0.138 & 0.103 & 0.137 & 0.033 & 0.182 & 0.277 & 0.164 & 0.175 & 0.300 & 0.138 & 0.054 & 0.138 & 0.028 & 0.102 & 0.033 & 0.006\\
			\hline                                        DEC & 0.257 & 0.301 & 0.161 & 0.136 & 0.185 & 0.050 & 0.276 & 0.359 & 0.186 & 0.282 & 0.381 & 0.203 & 0.122 & 0.195 & 0.079 & 0.115 & 0.037 & 0.007\\
			\hline                                        DAC & 0.396 & 0.522 & 0.306 & 0.185 & 0.238 & 0.088 & 0.366 & 0.470 & 0.257 & 0.394 & 0.527 & 0.302 & 0.219 & 0.275 & 0.111 & 0.190 & 0.066 & 0.017\\ 
			\hline                                        ADC & - & 0.325 & - & - & 0.160 & - & - & 0.530 & - & - & - & - & - & - & - & - & - & -\\
			\hline                                        DDC & 0.424 & 0.524 & 0.329 & - & - & - & 0.371  & 0.489 & 0.267 & 0.433 & 0.577 & 0.345 & - & - & - & - & - & -\\
			\hline                                        DCCM & 0.496 & 0.623 & 0.408 & 0.285 & 0.327 & 0.173 & 0.376 & 0.482 & 0.262 & 0.608 & 0.710 & 0.555 & 0.321 & \textbf{0.383} & 0.182 & 0.224 & 0.108 & 0.038\\
			\hline                                        IIC & - & 0.617 & - & - & 0.257 & - & - & 0.610 & - & - & - & - & - & - & - & - & - & -\\
			\hline                                        PICA & \textbf{0.591} & 0.696 & \textbf{0.512} &  \textbf{0.310} & 0.337 & 0.171 & 0.611 & 0.713 & 0.531 & 0.802 & 0.870 & 0.761 & 0.352 & 0.352 & 0.201 & 0.277 & 0.098 & 0.040\\
			\hline
			\hline                                        DCDC & 0.585 & \textbf{0.699} & 0.506 & \textbf{0.310} & \textbf{0.349} & \textbf{0.179} & \textbf{0.621} & \textbf{0.734} & \textbf{0.547} & \textbf{0.817} & \textbf{0.879} & \textbf{0.787} & \textbf{0.360} & 0.365 & \textbf{0.207} & \textbf{0.323} & \textbf{0.164} & \textbf{0.073}\\
			\hline
	\end{tabular}
	\label{results}
\end{table*}
\subsection{Model Training Settings}
The overall loss function (Eq.~\ref{total_loss}) is differentiable, scalable and elegant for standard SGD optimizer to train a end-to-end DNN model. At specific implementation, we randomly perturb the train data distribution (such as image random crop, image color jitter and image random grayscale) to do data augmentation for improving the robustness of the learned model. We strength the clustering results invariance over data perturbations from both sample and class views. Specifically, we obtain the original batch $U\in\mathbb{R}^{B\times C}$ and augmented one $U'\in\mathbb{R}^{B\times C}$ from original data and augmented one at each iteration. From the sample view of $U$ and $U'$, we obtain $u$ and $u'$ for Eq.\ref{sample_contra}. From the class view $U$ and $U'$, we obtain $v$ and $v'$ for Eq.\ref{class_contra}. The training method of DCDC is summarized in Algorithm \ref{algorithm1}. Note that for easy to tune the hyper-parameter, in our experiments, the temperature parameter $\tau$ is same for two contrastive losses.

\section{Experiments}
\subsection{Datasets}
In the experiments, we assess our DCDC on six widely used object recognition datasets:
\begin{enumerate}
	\item[\textbf{(1)}] \textbf{CIFAR-10/100} \cite{krizhevsky2009learning}: A natural image dataset with 50,000/10,000 samples from 10(/100) classes for training and testing respectively.
	\item[\textbf{(2)}] \textbf{STL-10} \cite{coates2011analysis}: An ImageNet \cite{russakovsky2015imagenet} sourced dataset containing 500/800 training/test images from each of 10 classes and additional 100,000 samples from several unknown categories.
	\item[\textbf{(3)}] \textbf{ImageNet-10 and  ImageNet-Dogs} \cite{chang2017deep}: Two subsets of ImageNet, the former with 10 random selected subjects and the latter with 15 kinds of dog.
	\item[\textbf{(4)}] \textbf{Tiny-ImageNet} \cite{le2015tiny}: A subset of ImageNet with 200 classes, which is a challenging dataset for clustering task. There are 100,000/10,000 training/test images evenly distributed in each category.
\end{enumerate}
We summarize the statistics of these datasets in Table~\ref{datasets_statistics}. For fair comparison, we utilize the same clustering setting as \cite{ji2019invariant,wu2019deep,chang2017deep,huang2020deep}: using both training and test sets (no labels) for CIFAR-10/100 and STL-10, and set the 20 super-classes as the ground truth of CIFAR-100.

\begin{table}[!h]
	\centering
	\caption{Statistics of six benchmark datasets.}
		\renewcommand\arraystretch{1.1}
	\begin{tabular}{c|ccc}
		\hline
		Datasets  & Classes & Total Images & Image Size \\ \hline
		CIFAR-10 & 10 & 60,000 & 32 $\times$ 32 $\times$ 3 \\
		CIFAR-100 & 20 & 60,000 & 32 $\times$ 32 $\times$ 3 \\
		STL-10  & 10 & 13,000 & 96 $\times$ 96 $\times$ 3 \\
		ImageNet-10 & 10 & 13,000 & 96 $\times$ 96 $\times$ 3 \\
		ImageNet-Dogs & 15 & 19,500 & 96 $\times$ 96 $\times$ 3 \\
		Tiny-ImageNet & 200 & 100,000 & 64 $\times$ 64 $\times$ 3 \\ \hline
	\end{tabular}
	\label{datasets_statistics}
\end{table}

\subsection{Evaluation Metrics}
There are three common clustering performance metrics in our experiments: Accuracy (\textbf{ACC}), Normalised Mutual Information (\textbf{NMI}) and Adjusted Rand Index (\textbf{ARI}).  
\begin{itemize}
	\item Accuracy (\textbf{ACC}) is computed by taking the average correct classification rate as the final score,
	\item Normalised Mutual Information (\textbf{NMI}) quantifies the normalised mutual dependence between the predicted labels and the ground truth, 
	\item Adjusted Rand Index (\textbf{ARI}) evaluates the predicted labels as a series of decisions and measures its quality according to how many positive/negative sample pairs are correctly assigned to the same/different clusters. 
\end{itemize} 
All these metrics scale from $ 0 $ to
$ 1 $ and higher values indicate better performance. The predicted label is assigned by the dominating class label.

\subsection{Implementation Details}
Our DCDC network is implemented by PyTorch 1.4 \cite{paszke2017automatic} and optimized by Adam \cite{kingma2014adam} with fixed learning rate $10^{-3}$. And we set a ResNet-like DNN as the backbone of our learning framework. Consistent with previous approaches \cite{caron2018deep,ji2019invariant,huang2020deep}, we still adopt the additional over-clustering head to increase expressivity of the learned feature representation. For the over-clustering head, we set 700 clusters for Tiny-ImageNet and 70 clusters for others. All models are randomly initialized and trained with 200 epochs. We set the batch size proportional to class numbers (due to various class numbers) and repeated each in-batch sample $r=3$ times. We summarize the specific hyper-parameters in Table~\ref{hy_statistics}. 

\begin{table}[h]
	\centering
	\caption{Hyper-parameters of six benchmark datasets.}
		\renewcommand\arraystretch{1.1}
	\begin{tabular}{c|cc}
		\hline
		Datasets  & Temperature $\tau$ & Batch Size $B$ \\ \hline
		CIFAR-10 & 0.5 & 50 \\
		CIFAR-100 & 0.3 & 200 \\
		STL-10  & 0.3 & 100 \\
		ImageNet-10 & 0.5 & 50 \\
		ImageNet-Dogs & 0.5 & 30 \\
		Tiny-ImageNet & 0.5 & 300 \\ \hline
	\end{tabular}
	\label{hy_statistics}
\end{table}

\subsection{Results about Comparison Methods and DCDC}
The following baselines will be compared against our DCDC network, including both conventional clustering methods and deep ones: K-Means \cite{macqueen1967some}, Spectral Clustering (SC) \cite{zelnik2005self}, Agglomerative Clustering (AC) \cite{gowda1978agglomerative}, Nonnegative Matrix Factorization (NMF) \cite{cai2009locality}, AutoEncoder (AE) \cite{bengio2007greedy}, Denoising AutoEncoders (DAE) \cite{vincent2010stacked}, Deep Convolutional Generative Adversarial Networks (DCGAN) \cite{radford2015unsupervised}, DeConvolutional Neural Networks (DeCNN) \cite{zeiler2010deconvolutional},  Variational Auto-Encoding (VAE) \cite{kingma2013auto}, Joint Unsupervised LEarning (JULE) \cite{yang2016joint}, Deep Embedding Clustering (DEC) \cite{xie2016unsupervised}, Deep Adaptive Clustering (DAC) \cite{chang2017deep}, Associative Deep Clustering (ADC) \cite{haeusser2018associative}, Deep Discriminative Clustering (DDC) \cite{chang2019deep}, Deep Comprehensive Correlation Mining (DCCM) \cite{wu2019deep}, Invariant Information Clustering (IIC) \cite{ji2019invariant}, PartItion Confidence mAximisation (PICA) \cite{huang2020deep}. The results of previous methods are taken from \cite{wu2019deep,ji2019invariant,huang2020deep}, and finally are summarized in Table~\ref{results}. From the results, our DCDC method always superior than other SOTA methods, which demonstrate the effectiveness of our method. Particularly in CIFAR-100, STL-10 and Tiny-ImageNet datasets, DCDC leads 1.2\%, 2.1\%, 5.6\% against other comparison methods. Although in ImageNet-Dogs dataset, the ACC results of DCDC lower than DCCM method (NMI and ARI higher), but it still head of PICA method in three evaluation metrics.

\subsection{Ablation Study}
In this section, we conduct several ablation studies to demonstrate the effect of different choices in DCDC.
\subsubsection{Effect of doubly contrastive loss:} 
At first, we assess the effect of two contrastive losses respectively and present the results in Table~\ref{ablation_loss}. From the Table, we can make a statement that the sample contrastive loss is more effective than the class one. The reason is that our experiments adopt batch size proportional to class numbers. For larger dataset such as Tiny-ImageNet, its batch size is set as 300, while the class number is 200, it is more likely that one class have at least one sample. In this situation, sample contrastive loss can dominate the optimization procedure which leads to better performance.
\begin{table}[h]
	\centering	
	\caption{Effect of doubly contrastive loss. Metric: ACC.}
		\renewcommand\arraystretch{1.1}
	\setlength{\tabcolsep}{0.5mm}
	\begin{tabular}{c|ccc}
		\hline
		& CIFAR-10 & CIFAR-100 & Tiny-ImageNet\\ \hline
		Sample Contrastive & 0.606  & 0.307 & 0.098\\
		Class Contrastive & 0.458 & 0.209 & 0.092\\ 
		DCDC  & \textbf{0.699} & \textbf{0.349} & \textbf{0.164}   \\
		\hline
	\end{tabular}
	\label{ablation_loss}
\end{table}

\subsubsection{Effect of over-clustering:}
According to previous approaches \cite{caron2018deep,ji2019invariant,huang2020deep}, the utilization of over-clustering head have two purposes: 1. Making use of extra noisy data with distractor classes to mine more information (such as, STL-10). 2. For the no extra data case, increasing expressivity of the learned feature representation (such as CIFAR-100 and Tiny-ImageNet). The results are presented at Table~\ref{ablation_over}. Obviously, over-clustering technique is helpful to improve performance in both two cases. Particularly in STL-10 dataset, over clustering technique improves a lot performance.
\begin{table}[h]
	\centering
	\caption{Effect of over-clustering. Metric: ACC.}
		\renewcommand\arraystretch{1.1}
	\setlength{\tabcolsep}{1mm}	
	\begin{tabular}{c|ccc}
		\hline
		& STL-10 & CIFAR-100 & Tiny-ImageNet\\ \hline
		No Over-clustering & 0.612  & 0.298 & 0.159\\
		DCDC & \textbf{0.734} & \textbf{0.349} & \textbf{0.164}   \\
		\hline
	\end{tabular}
	\label{ablation_over}
\end{table}

\subsubsection{Effect of sample repeat:}
For data augmentation, we repeat samples within each batch $r$ times; this means that multiple sample pairs within a batch contain the same original image and its different transformation version, which encourages greater information mining since there are more samples of which visual details to ignore \cite{ji2019invariant}. The results are provided in Table~\ref{ablation_repeat}. As shown in the results, sample repeat strategy can improve the performance in a certain degree.
\begin{table}[h]
	\centering
	\caption{Effect of sample repeat. Metric: ACC.}
		\renewcommand\arraystretch{1.1}
	\setlength{\tabcolsep}{3.2mm}
	\begin{tabular}{c|cc}
		\hline
		& CIFAR-100 & Tiny-ImageNet\\ \hline
		No Sample Repeat & 0.328 & 0.163\\
		DCDC & \textbf{0.349} & \textbf{0.164}   \\
		\hline
	\end{tabular}
	\label{ablation_repeat}
\end{table}

\subsubsection{Effect of batch size:}
According to current contrastive learning methods \cite{cheng2020club}, we can easily make a statement that large batch size can achieve better performance. But this statement may not suitable for deep clustering task. Since contrastive learning methods always adopt ImageNet dataset which have 1,000 classes to learn feature representation, it is necessary to adopt large batch size beyond 1,000 to make sure that each class have at least one sample. However, in deep clustering, the largest class number is Tiny-ImageNet, only 200 classes, therefore, we adopt 300 batch size and then obtain more performance improvement. It seems that simply set the batch size close to class number may obtain better performance. However, considering the smaller batch size leads to more time consumption, we can not simply do it. That is also the reason why we adopt the batch size proportional to class numbers. 

Note that, this conclusion does not conflict with the statement about lower bound (Eq.\ref{lower_bound}) that the more $N$, the lower $\mathcal{L}_{InfoNCE}$. The precondition of this statement is that we should keep one positive sample and $N-1$ negative samples. However, for the smaller class number with large batch size cases, this precondition not holds any more. The results on CIFAR-100 are shown in Table~\ref{ablation_batch}. A proper batch size helps to improve clustering performance.
\begin{table}[h]
	\centering
		\renewcommand\arraystretch{1.1}
	\caption{Effect of batch size on CIFAR-100 dataset.}
	\setlength{\tabcolsep}{5.5mm}
	\begin{tabular}{c|ccc}
		\hline
		Batch Size & NMI & ACC & ARI \\ \hline
		100 & 0.267 & 0.296 & 0.142 \\
		200 & 0.310 & \textbf{0.349} & \textbf{0.179}  \\
		300 & \textbf{0.316} & 0.342 & \textbf{0.179}  \\
		400 & 0.305 & 0.323 & 0.165  \\
		500 & 0.305 & 0.325 & 0.168  \\
		1000 & 0.290 & 0.302 & 0.155  \\
		\hline
	\end{tabular}
	\label{ablation_batch}
\end{table}

\subsection{Qualitative Study}
\subsubsection{$t$-SNE Visualization:}
\begin{figure}[!h]
	\centering	
	\subfloat[PICA]{
		\centering
		\includegraphics[width=0.22\textwidth]{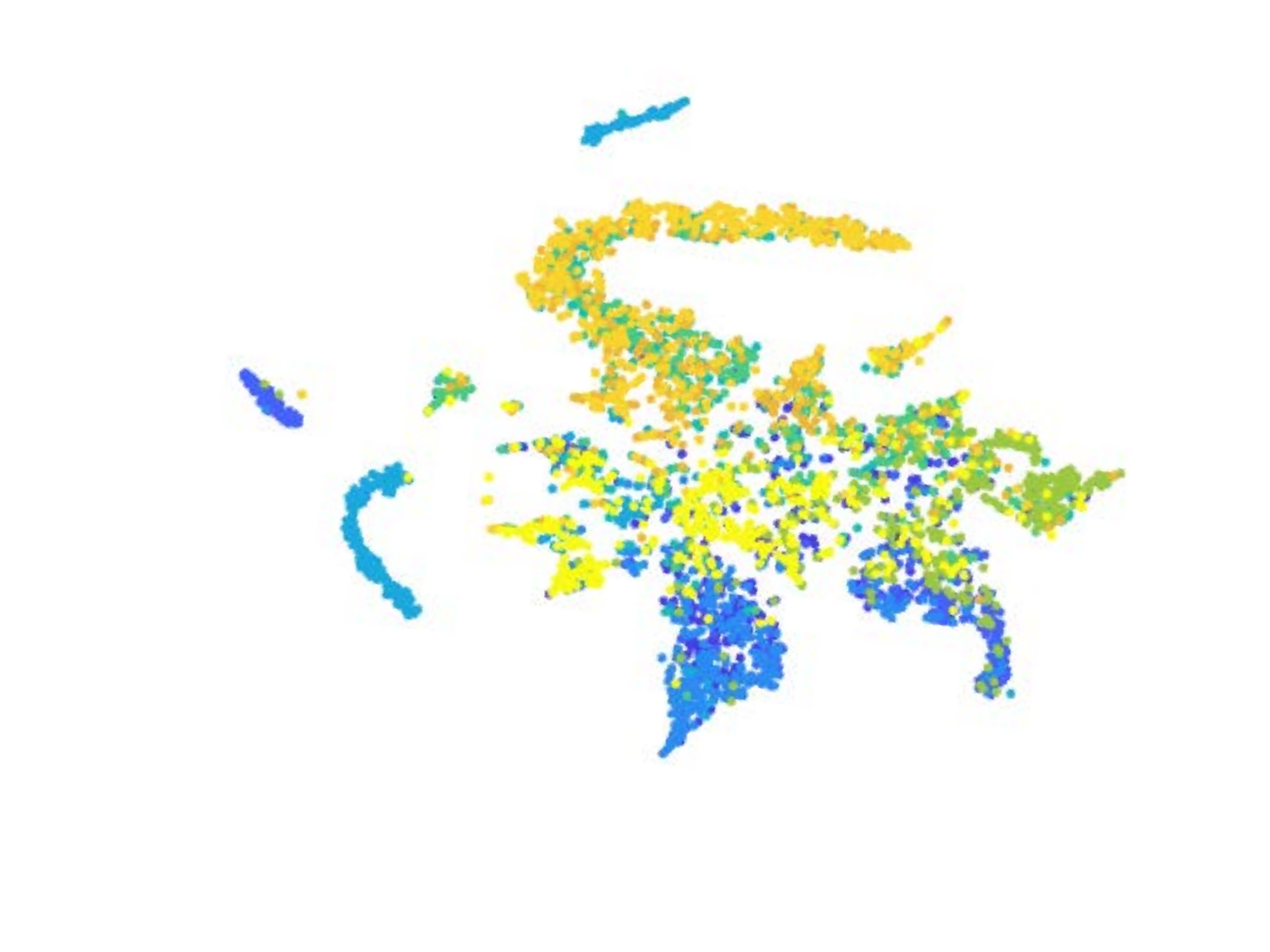}
		\label{tsne_pica}
	}
	\subfloat[DCDC]{
		\centering
		\includegraphics[width=0.22\textwidth]{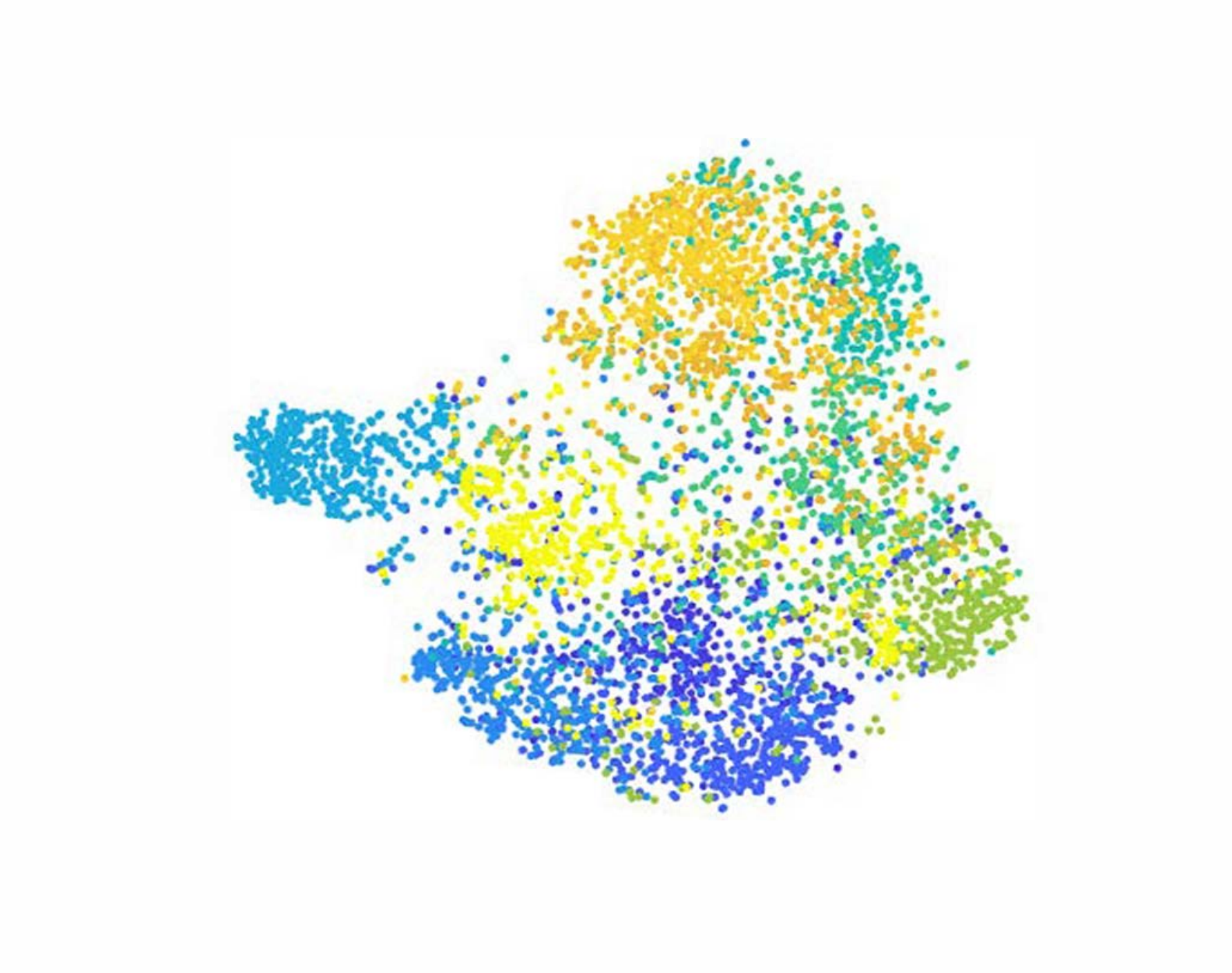}
		\label{tsne_dc}
	}
	\caption{$t$-SNE visualization to show the discriminative capability of the learned features on the former 10 classes of Tiny-ImageNet dataset.}
	\label{tsne_tiny}
\end{figure}

We adopt $t$-SNE visualization \cite{maaten2008visualizing} technique to show the discriminative capability of the learned features on Tiny-ImageNet dataset which is presented in Figure~\ref{tsne_tiny}. From the results, although PICA have some separate features, there are more features mix up at the center of the Figure~\ref{tsne_pica}. On the contrary, due to novel introduced sample-wise contrastive loss, our method achieves more congregate features than PICA. That is the reason why DCDC obtains better performance.

\subsubsection{Desired Affinity Matrices:}
Recall the description in Figure~\ref{framework}: sample-wise affinity matrix $M$ will have block-diagonal structure (if data are sort by class) and class-wise affinity matrix $N$ still is diagonal structure, Figures~\ref{illustration_sample} and \ref{illustration_class} provide the visible presentations of two affinity matrices which prove the correctness of this statement.

\begin{figure}[!t]
	\centering	
	\subfloat[PICA]{
		\centering
		\includegraphics[width=0.22\textwidth]{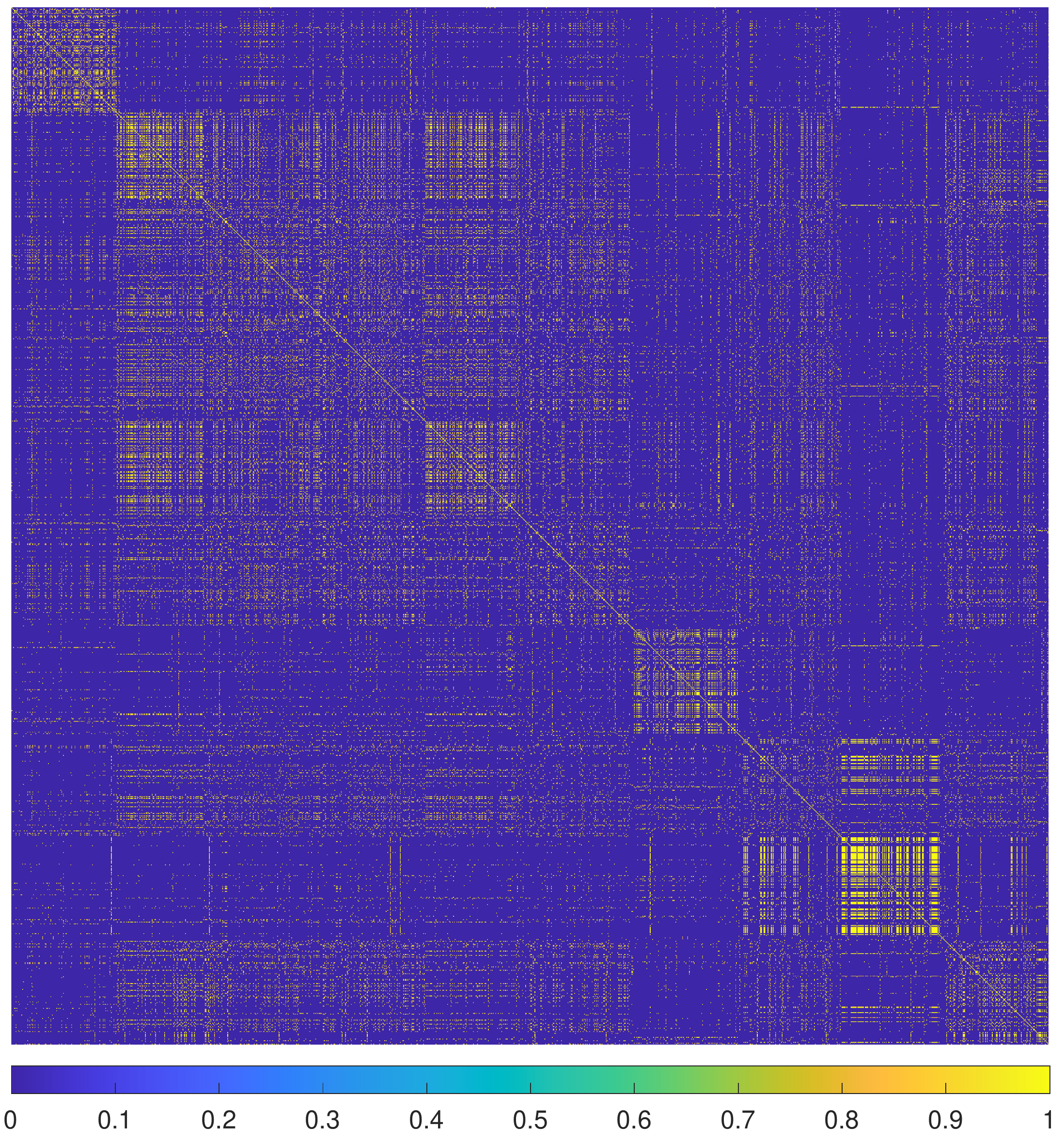}
		\label{sample_pica}
	}
	\subfloat[DCDC]{
		\centering
		\includegraphics[width=0.22\textwidth]{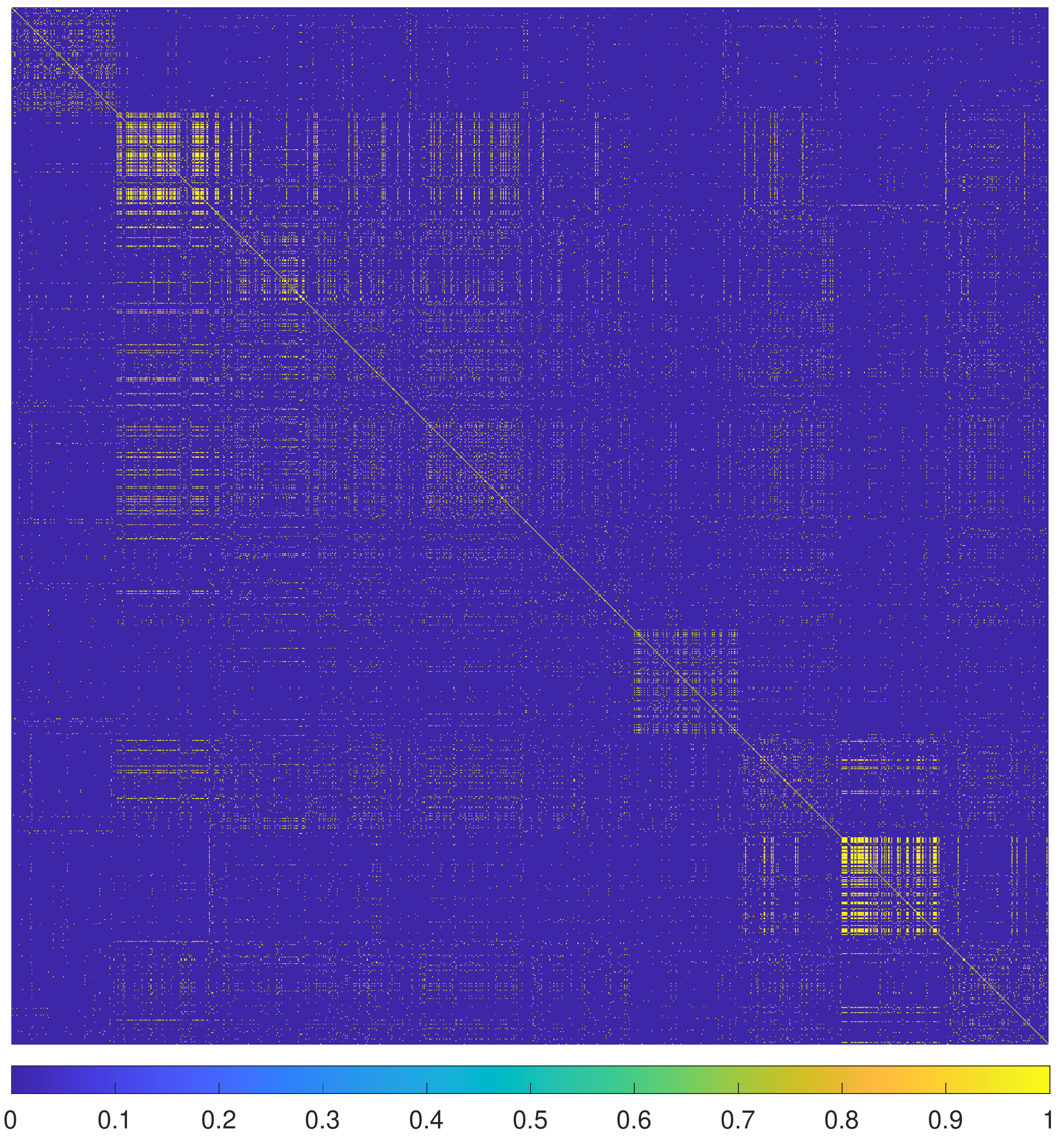}
		\label{sample_dc}
	}
	\caption{Sample-wise affinity matrix ($M$ in Figure~\ref{framework}) comparison of two methods on the former 10 classes of Tiny-ImageNet dataset.}
	\label{sample_affinity_tiny}
\end{figure}

\begin{figure}[!t]
	\centering
	\subfloat[PICA]{
		\centering
		\includegraphics[width=0.22\textwidth]{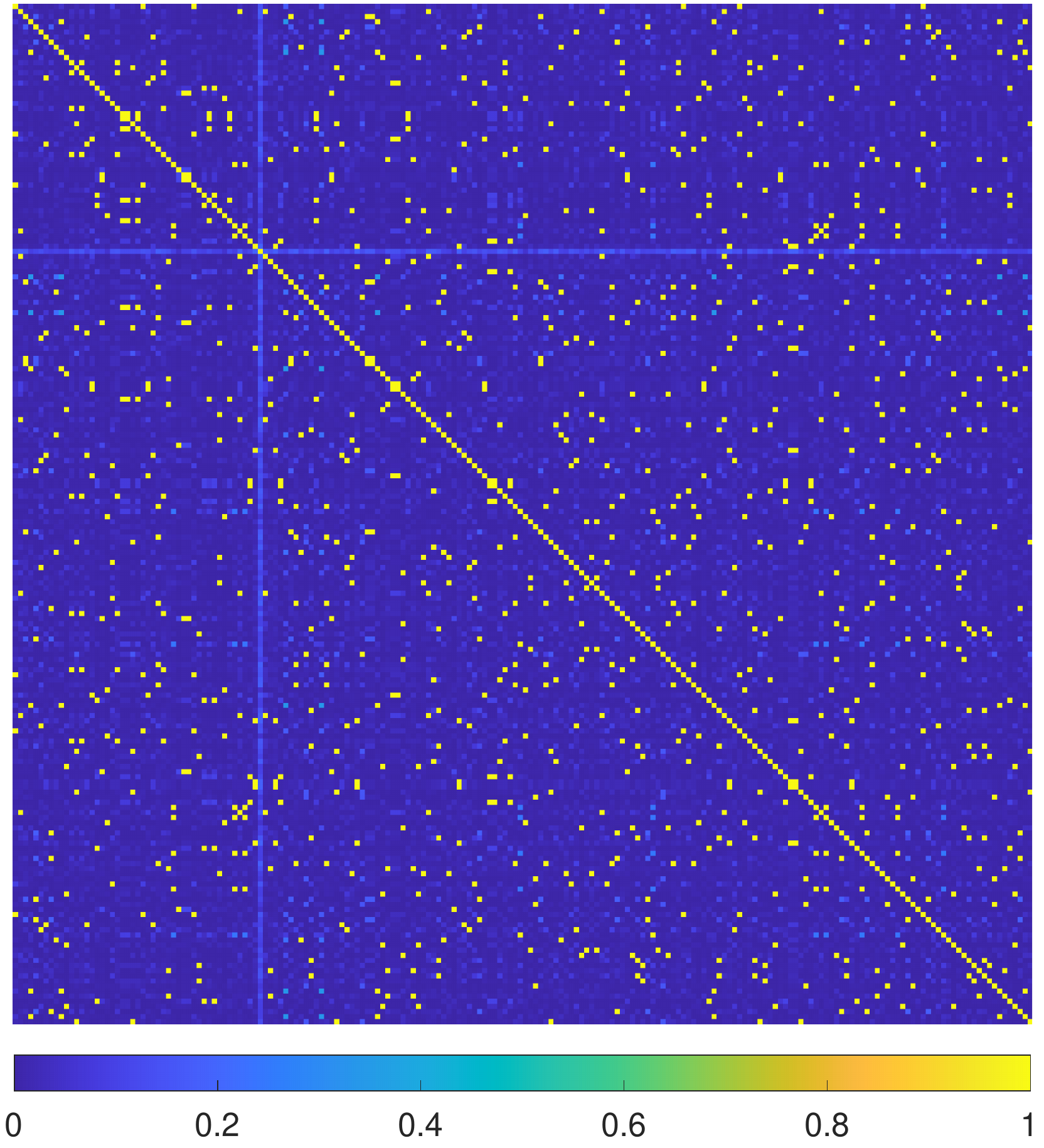}
		\label{class_pica}
	}
	\subfloat[DCDC]{
		\centering
		\includegraphics[width=0.22\textwidth]{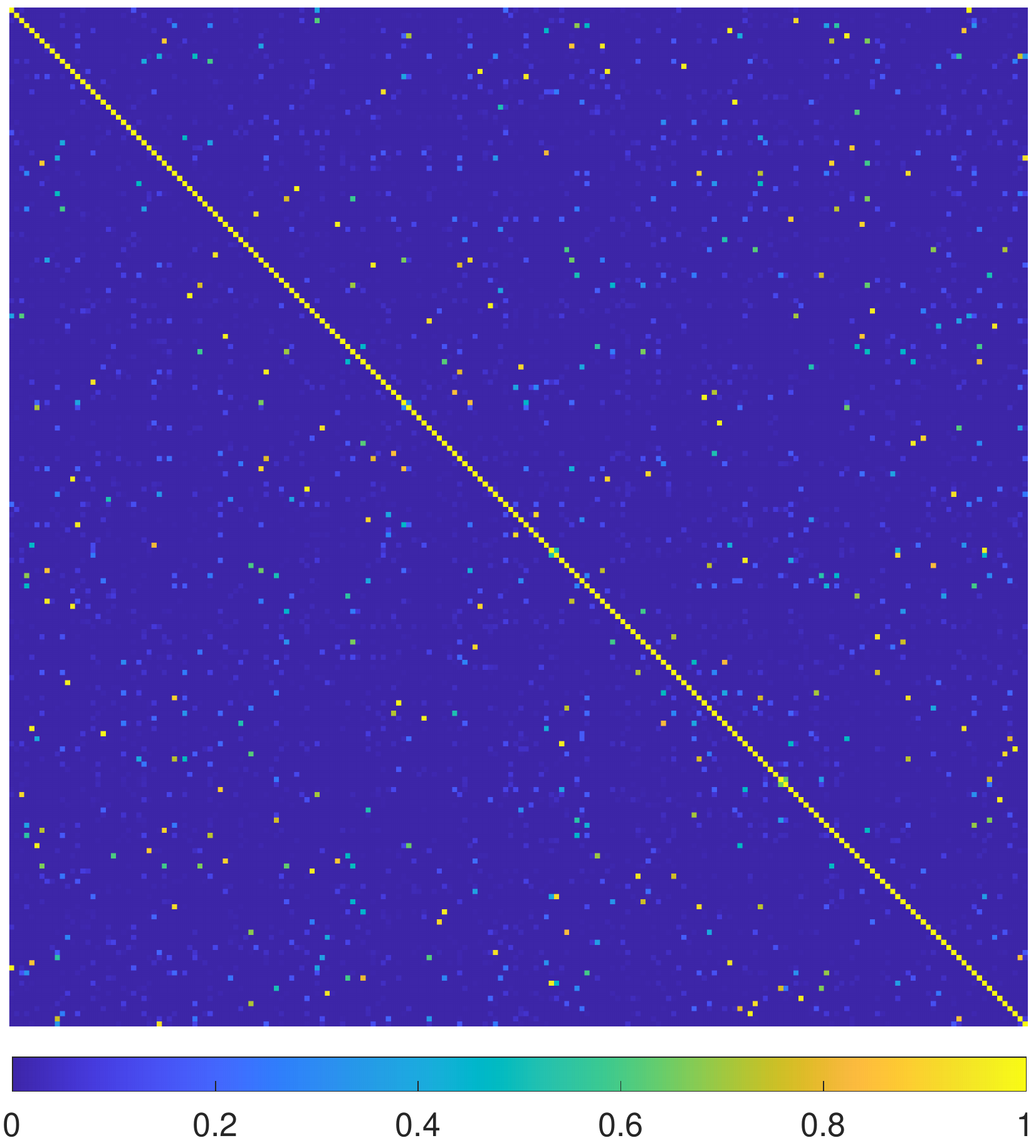}
		\label{class_dc}
	}
	\caption{Class-wise affinity matrix ($N$ in Figure~\ref{framework}) comparison of two methods on Tiny-ImageNet dataset.}
	\label{class_affinity_tiny}
\end{figure}

Since $M=cos(UU'^T)$ and $N=cos(U^TU')$, the cosine similarity value scale from $0$ to $1$, higher value indicate the point will be more yellow in the Figures. In Figure~\ref{sample_affinity_tiny}, our DCDC has less perturbation points than PICA method which means sample-wise contrastive loss is helpful to obtain more reasonable class distributions of samples. Figure~\ref{class_affinity_tiny} also presents the same situation which indicates the effectiveness of class-wise contrastive loss.

\section{Conclusion}
We have proposed a novel deep clustering framework based two contrastive losses, named Doubly Contrastive Deep Clustering (DCDC). Different from previous methods, DCDC loss considers improving clustering performance from both sample and class views. Benefiting from this novel loss, our network on six benchmark datasets outperforms state-of-the-art methods. Besides that, we also conduct more ablation studies about over-clustering, sample repeat and batch size to figure out the useful techniques or tricks for contrastive based deep clustering. 

\bibliography{bare_jrnl}
\bibliographystyle{IEEEtran}

\end{document}